\pdfoutput=1

\documentclass[11pt]{article}
\usepackage{setspace}
\usepackage[dvipsnames]{xcolor}
\usepackage{graphicx}
\usepackage{tabularx}
\usepackage{afterpage}
\usepackage{booktabs}
\usepackage{amssymb,amsmath}
\usepackage{subcaption}
\usepackage{xspace}
\usepackage{etoolbox}
\usepackage{enumitem} %
\usepackage{cellspace} %
\usepackage{acl} 
\providetoggle{showcomments}
\toggletrue{showcomments}

\usepackage{times}
\usepackage{latexsym}
\usepackage{xspace}
\usepackage[T1]{fontenc}

\usepackage[utf8]{inputenc}

\usepackage{microtype}

\usepackage{inconsolata}
\usepackage{float}

\title{
Taking a turn for the better: \\ Conversation redirection throughout the course of \mht
}

\newcommand{\nstwo}{\hspace*{.2in}}
\newcommand{\nsone}{\hspace*{.1in} }%
\author{~ \\
\cornellian{Vivian Nguyen}\nsone \cornellian{Sang Min Jung}\nsone\cornellian{Lillian Lee} \\
\textbf{\talkspacer{Thomas D. Hull}\nsone   \cornellian{Cristian Danescu-Niculescu-Mizil}\thanks{Senior corresponding author.}}\\
vn72@cornell.edu\nsone sj597@cornell.edu\nsone {llee@cs.cornell.edu} \\
{derrick@talkspace.com}\nstwo {cristian@cs.cornell.edu} \\
 $\textcolor{white}{~}^\cornellmarker$ Cornell University\nstwo  $\textcolor{white}{~}^\talkspacemarker$Talkspace \\
}

\newcommand{\cut}[1]{}
\newcommand{\xhdr}[1]{{\noindent\bfseries #1.}}

\newcommand{\redirection}{redirection\xspace}
\newcommand{\Redirection}{Redirection\xspace}
\newcommand{\RedirectionBoldLetter}{{\bf R}edirection\xspace}
\newcommand{\redirections}{redirections\xspace}

\newcommand{\redirecting}{redirecting\xspace}
\newcommand{\redirect}{redirect\xspace}
\newcommand{\redirects}{redirects\xspace}

\newcommand{\mh}{mental-health\xspace}
\newcommand{\Mh}{Mental-health\xspace}
\newcommand{\mht}{\mh therapy\xspace}
\newcommand{\Mht}{\Mh therapy\xspace}

\newcommand{\definedas}{\overset{\triangle}{=}}

\newcommand{\cornellmarker}{{\ensuremath{\color{red}{\bullet}}}}
\newcommand{\cornellian}[1]{\bf $\mbox{#1}^\cornellmarker$}
\newcommand{\talkspacemarker}{{\rm TS}}
\newcommand{\talkspacer}[1]{\bf $\mbox{#1}^\talkspacemarker$}

\newcommand{\contrib}[1]{{{\bf #1}}}

\definecolor{therapist}{RGB}{36,97,189}
\definecolor{client}{RGB}{244,130,130}

\begin{document}

\maketitle

\begin{abstract}

\Mht involves a complex conversation flow in which patients and therapists continuously negotiate what should be talked about next.
For example, therapists might try to shift the conversation's direction to keep the therapeutic process on track and avoid stagnation, or patients might push the discussion towards issues they want to focus on.

How do such patient and therapist \redirections relate to the development and quality of their relationship?
To answer this question, we introduce a probabilistic measure 
of the extent to which a certain utterance immediately \textit{\redirects} the flow of the conversation,
accounting for both the intention and the actual realization of such a change.
We apply this new measure to characterize the development of patient-therapist relationships over multiple sessions in a very large, widely-used online therapy platform. 
Our analysis reveals that (1) patient control of the conversation's direction generally increases relative to that of the therapist as their relationship progresses; 
and (2) patients who have less control in the first few sessions are significantly more likely to eventually express dissatisfaction with their therapist and terminate the relationship.

\end{abstract}

\section{Introduction}
\label{sec:intro}

\Mht conversations are remarkably and consequentially complex.
They involve an ongoing negotiation between a patient and a therapist
regarding what should be talked about, how it should be talked about, and by whom. 
Conversational bids to shift the direction of the discussion (henceforth \textit{\redirections}) are a common and essential aspect of any therapeutic relationship.
However, we lack an understanding of how these dynamics play out throughout the course of the therapeutic relationship and how they might relate to its quality.

\begin{figure}[t]
    \centering
    \includegraphics[width=1\linewidth]{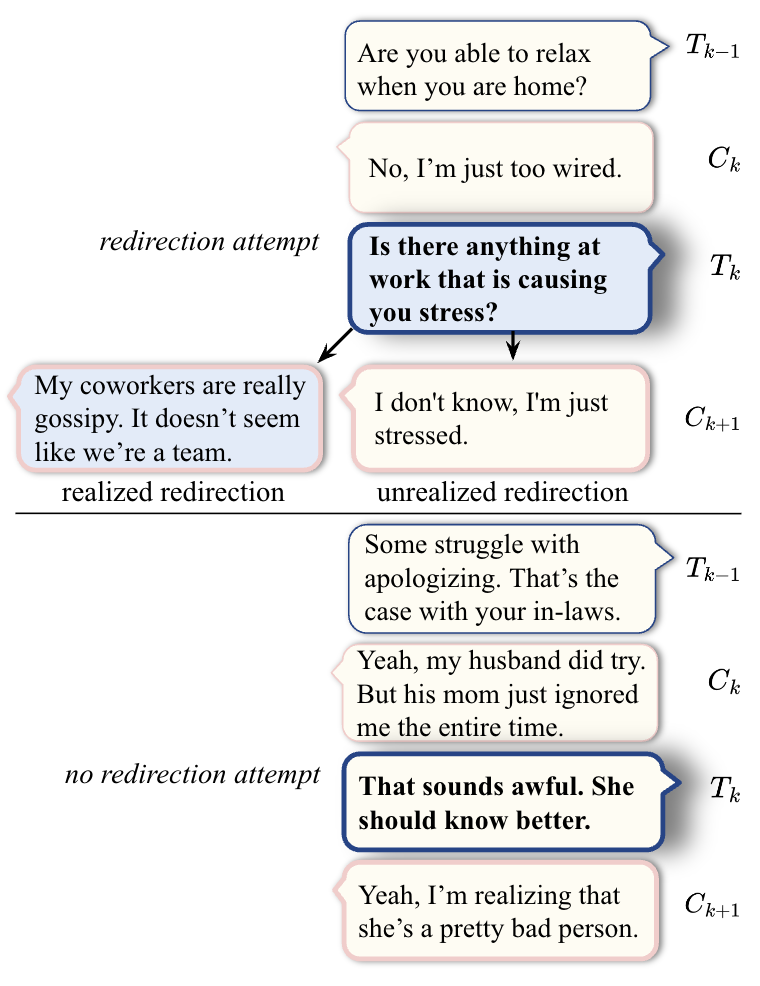}
    \caption{
    \textbf{Top:} Examples of attempted redirection, both realized (left) and unrealized (right).
    \textbf{Bottom:} Example 
    where redirection is not attempted.
    $T_k$ and $C_k$ refer to 
    the therapist's and patient's $k^{\rm th}$ utterance, respectively. Note that in general, both parties can \redirect.
    }
    \label{fig:intro_example}
\end{figure}

One key aspect of {\redirections} that may explain why they have yet to be studied in a systematic and rigorous way is their joint, rather single-utterance, manifestation.
In contrast to other commonly studied strategies and conversational acts \cite{malgaroli_natural_2023} that tend to be executed by either the therapist (e.g., empathy \cite{sharma_computational_2020}) or the patient (e.g., change talk \cite{park_conversation_2019}), \redirection is a truly 
shared patient-therapist act: to be realized, a bid to redirect the conversation must be accepted by the other participant.

Our \contrib{first contribution} is to
introduce a measure to quantify an utterance's \redirection effect in a way that explicitly takes into account this 
two-part
nature: it jointly captures
both
the \textit{intention} to change the course of the conversation and the actual \textit{realization} of this change through a reply that complies
with this intention.
\autoref{fig:intro_example} (top, left branch) illustrates an example of an utterance with high \redirection effect:  the therapist intends to switch the focus towards specific sources of stress, and the patient conforms by mentioning frustration at work. 
In contrast, if the patient resists the switch (\autoref{fig:intro_example}, top, right branch), the \redirection is not realized.  
There is a third alternative: \autoref{fig:intro_example} (bottom) shows an utterance with low \redirection effect due to the lack of intention to \redirect.

Key to our work is capturing this type of interplay between two participants at a specific utterance juncture,  rather than a discourse-level notion of shift, where  a single person can change the focus even on their own (e.g., topic control \cite[inter alia]{nguyen_modeling_2014}).
Our \redirection measure's inherent joint nature is particularly well-suited for examining the therapeutic relationship, and thus adds to the toolkit of computational methods available for studying therapeutic practice \cite{althoff_large-scale_2016,zhang_balancing_2020, imel_outcomes_2024, yang_modeling_2024}.
We illustrate this in collaboration with Talkspace---a large online text-based therapy platform---by applying it to characterize the development and perceived quality of the patient-therapist relationship.

Our analysis reveals that as their relationship develops, both patients and therapists redirect to a lesser extent, suggesting an overall smoother and more focused conversation flow.  
However, in relative terms, patients gain increasing control
of the direction of the conversation in comparison to the therapists; \contrib{this discovery is our second contribution}.
Furthermore, as \contrib{our third contribution}, we find that patients are more likely to eventually express dissatisfaction with the therapeutic relationship and terminate it when they are not able to \redirect the conversation in the first few sessions. 
Cumulatively, these results underline the importance of considering patient agency in psychotherapy \cite{carey_will_2010,huber_agency_2021}, or, concomitantly, therapist willingness to follow the patient's lead.

While here we focus on the \mh domain, 
{\bf one final contribution} is to highlight the possibilities in examining any sort of longer-term relationships developed through multiple conversations over time.   Future work could apply 
our measure and analysis framework to study the conversational process in other conversation-rich settings where complex long-term relationships are developed, such as education (e.g., advisor-advisee relationships).  
To encourage such work, we make our code publicly available as part of ConvoKit \cite{chang_convokit_2020}, together with a demo on a publicly-available dataset in a different domain, US Supreme Court oral arguments (\autoref{sec:supreme}).\footnote{\url{https://convokit.cornell.edu/}}

\section{Therapy Setting}
\label{sec:data}

\xhdr{Long-term text-based therapy}
We develop our methodology in the context of Talkspace,
a telehealth 
therapy
platform.
Patients can choose between different plans, which may include video 
therapy alongside messaging therapy. 
After a consultation in which they answer a few questions about their symptoms and describe their preferences, patients are matched by the platform to a suitable therapist who is licensed in their state.

In this work, we use the (English-language) text-based 
conversations
(after redaction of personally identifiable information by Talkspace) from a five-year span.\footnote{The use of the therapist and patient data is done with their consent, and this research was approved by the Cornell IRB.}
During this time, Talkspace hosted over 18,000 licensed therapists providing services to over 300,000 patients.
In total, over 65 million 
messages were exchanged.

Therapies can be sustained over long periods of time, with 
more than 26,000 therapies lasting over a year and 17,000 therapies comprising over 500 messages.
This setting thus provides an opportunity to study the long-term conversational
dynamics as the therapeutic relationship develops.

At any point, patients can either cancel their therapy or switch to a different therapist.  
When doing so, they are asked to provide a reason, either by selecting from a drop-down menu (e.g., ``The treatment provided by my therapist was not helpful,'' ``I met my goal / I feel better'') or by entering free text.
We have some assurance that these reasons are authentic because patients are informed that therapists cannot access them.
These reasons serve as imperfect indicators of the perceived quality of the therapeutic relation (for a broader discussion of difficulties in operationalizing the quality of the therapeutic 
relationship, see Section~\ref{sec:limitations}).

\xhdr{Session identification}
One challenge in studying the development of therapeutic relationships in this text-based setting is accounting for the wide variety of durations, tempos, and volubility exhibited in different therapies.
A longitudinal analysis requires a methodology that captures the progression of the therapeutic process while allowing for meaningful aggregation across therapies with these vastly different interaction patterns.

To address this challenge, we need to account for several factors.
First, each therapist-patient pair develops a unique interaction style and tempo, including frequency of interaction, response time, and how often turns are split into multiple messages.
Second, exchanges between a therapist and a patient include not only therapeutic conversation, but also short check-ins that deal with brief updates, scheduling, or survey completions.
Last, an important component of many therapies is the occurrence of spans of {\em approximately synchronous} conversations, or \textit{sessions}, over the therapy's duration.
These three factors render conventional units of progression in conversational analysis---such as time or number of utterances---unsuited for this complex interactional setting.

Explicit session boundaries are lacking in text-based conversations and so must be inferred.
We start by distinguishing high-activity periods interwoven with periods of little or no activity, using the method from \citet{kushner_bursts_2020} to capture these bursts.
For each therapy, we
define session-split points as moments when an utterance's reply time 
exceeds $N \times$ (\textit{median reply time of that entire therapy}).\footnote{
We settle on $N=100$; see Appendix \ref{sec:appendixsessionsplit} for details.}
We further ensure that each session represents an actual conversation between the patient and the therapist (as opposed to, for example, a short notification and acknowledgement regarding scheduling) 
by filtering out short exchanges---any burst of activity with fewer than four turns (so that what remains includes at least two nonconsecutive utterances from each speaker)---or includes a video session.
We further remove any automated messages related to video sessions, survey completion, or scheduling.

To validate the quality of the session splits,
we sampled therapies with at least 10 sessions and ran 
a Bayesian distinguishing-word analysis \cite{monroe_fightin_2017} on the first and last utterance of each session.
The distinguishing words are intuitive: sessions start with greetings ("hi", "morning", "hey", "how") and end with a farewell or expression of gratitude ("enjoy", "welcome", "thank", "thanks").

\section{Redirection Measure}
\label{sec:method}
\subsection{What redirection is (not)}
We aim to quantify the \redirection effect of a given utterance in a conversation: the extent to which it alters the immediate focus of the conversation.
We start from the realization that for an utterance to have a high \redirection effect, it must (a) attempt to put the conversation on a different course (intention of redirection), and (b) receive a reply that is compliant with this redirection (realization of redirection). 
We will examine previous methods that capture these two conditions separately and explore how they guide us toward a design for a measure that jointly accounts for both of them.

For concreteness, we employ the convention (and notation) from \autoref{fig:intro_example} of centering the discussion on a therapist's utterance $T_k$; but the analogous definitions apply to a patient's utterance $C_k$.

\xhdr{Orientation}
Orientation \cite{zhang_balancing_2020} captures the degree to which an utterance \textit{intends} to move the conversation away from what was already discussed. 
A "forwards-oriented" utterance (high orientation) is one that intends to advance the conversation towards a specific target, and thus is \textit{expected} to be followed by a reply centered on that target. 
A "backwards-oriented" utterance (low orientation) is one that intends to address what was previously said, and thus is \textit{expected} to follow a specific utterance.
Thus, orientation characterizes an utterance's intended objective, regardless of whether that intention is realized or not.

In \autoref{fig:intro_example} (top), $T_k$ is a high-orientation utterance since the therapist intends to redirect the conversation by asking the patient about a specific source of stress at work. 
This remains true regardless of whether the patient conforms in $C_{k+1}$ by aligning themselves with the newly suggested direction (by introducing their frustration with gossip; left branch), or resists the redirection attempt (by maintaining the focus on their general state of stress, rather than delving into specifics; right branch).

\xhdr {Similarity difference}
To incorporate the actual reply, one may simply compute the similarity of the reply $C_{k+1}$ with the original utterance $T_{k}$ and compare it against a reference point to account for the ongoing direction of the conversation. 
A potential reference point is the extreme scenario when the therapist wishes to ensure no redirection will take place by simply repeating their previous utterance $T_{k-1}$. 
In other words, the \redirection of utterance $T_k$ could be formalized as the difference between the similarity of $C_{k+1}$ and $T_k$ and that of $C_{k+1}$ and $T_{k-1}$.

Unlike orientation, this measure considers the actual reply;
but it  hinges 
on the assumption that semantic similarity can sufficiently capture the \redirection realization. 
Successful redirection, however, is not synonymous with similarity between
utterance and reply. 
Two examples in \autoref{fig:intro_example} (top, left vs. right branch) illustrate this:  while they intuitively show different levels of redirection, in both cases the similarity difference is high.

\xhdr{Uptake}
For a more nuanced take on redirection realization that goes beyond similarity, we could aim to measure the reply's ``uptake'': how much it answers, acknowledges, or builds upon the previous utterance.
\citet{demszky_measuring_2021} formalize this concept (in the context of student-teacher interactions) as the dependence of the reply on the utterance via point-wise Jensen-Shannon divergence, which uses next-sentence \textit{prediction} to quantify the extent to which a given reply is a \textit{probable} response.

While capturing \redirection realization better than similarity, uptake does not account for the \redirection intent.
For instance, two examples in \autoref{fig:intro_example} (top left branch vs. bottom) have high uptake, with the patient replies being consistent with the respective therapist's utterance. 
They, however, show different levels of \redirection, as only in the first example does the therapist intentionally shift the focus of conversation.

\xhdr{Our approach: \Redirection}
We combine the insights of these existing measures to develop a new \redirection measure that captures both the intention and realization components.
In particular, our analysis of the previous measures points toward the need to consider the actual reply the utterance receives, to use a predictive component to capture the extent to which the reply naturally follows the utterance, and to use a point of reference to capture a change in direction.

More concretely, to quantify the \redirecting effect of an utterance, we first consider the likelihood of the patient’s reply given the previous context: 
\newcommand{\nextGivenTrue}{{\cal P}_{k}(C_{k+1})}
\newcommand{\nextGivenAlt}{{\cal Q}_{k}(C_{k+1})}
\begin{equation*}
\nextGivenTrue \definedas P(C_{k+1}|C_k,T_k).
\end{equation*}
\noindent We condition on the most recent utterances from both speakers, $C_k$ and $T_k$, to capture both of their prior conversational context.

We use as point of reference the extreme scenario wherein the therapist simply repeats their previous utterance $T_{k-1}$ as $T_k$, demonstrating absolutely zero intent to \redirect:\footnote{
One naive alternative might be to calculate $\nextGivenAlt$ without conditioning on the previous therapist utterance $T_{k-1}$, to simulate the ``absence'' of any therapist utterance. 
This alternative, however, creates an unnatural situation with the patient replying to themselves (which would artificially have a low probability). 
Furthermore, such an alternative would not be directly comparable with $\nextGivenTrue$, which is conditioned on the previous utterances of both speakers.
Therefore, we employ a reference point with equivalent conditioning.}
\begin{equation*}
\nextGivenAlt \definedas P(C_{k+1} | C_k, T_{\mathbf{k-1}}).
\end{equation*}

We then formally define the {\em \redirection} of $T_k$ as the log-odds ratio of these two probabilities:
\begin{equation}
\label{eq:redirection}
\scalebox{1}{$R(T_k) \definedas \log \left(\frac{\nextGivenTrue}{1 - \nextGivenTrue} \middle/ \frac{\nextGivenAlt}{1 - \nextGivenAlt} \right)$}
\end{equation}

When \redirection is high, 
$\nextGivenTrue \gg \nextGivenAlt$,
$T_k$ alters the direction of the conversation, in the sense that the patient’s reply $C_{k+1}$ is likely as a reply to this utterance, but not as a continuation of the previous direction of the conversation.
Conversely, in cases of low \redirection, the patient's response is less affected by the therapist's utterance.

We compute the \redirection of a patient’s utterance in the corresponding way.

\xhdr{Operationalization}
To compute the probabilities for our \redirection measure, we fine-tuned the Gemma-2B model \cite{mesnard_gemma_2024} with 4-bit QLoRA \cite{dettmers_qlora_2023} using a held-out dataset of $8,000$ therapy conversations.\footnote{The models were trained on internal GPU servers in consideration of the sensitive nature of the data. Training details are included in Appendix \ref{sec:appendiximplementation}.}

The operationalization for the related measures discussed above is included in Appendix~\ref{sec:appendiximplementation}.

\subsection{Session-level aggregation}
To apply and compare our utterance-level measures across therapy sessions, we use two aggregation methods.
The first method averages the utterance-level measures per session for each speaker. 
A speaker's average \redirection in a session is computed as the mean \redirection of their utterances in that session, which we denote with  $T_{avg}$ and  $C_{avg}$ for the therapist and patient respectively.

The second method examines the balance of \redirection between speakers to determine who is \redirecting the conversation more in a given session.
Analyzing the balance of \redirection can reveal how the control of the discussion is shared between the patient and therapist.
Equation \ref{eq:rel_redirection} formalizes relative \redirection from the therapist's perspective:
\begin{equation}
\label{eq:rel_redirection}
    T_{rel} = \frac{\exp(T_{avg})} {\exp(T_{avg}) + \exp(C_{avg})}.\
\end{equation}
The relative value for the patient $C_{rel}$ is computed the corresponding fashion.
Note that mathematically,
$T_{rel} + C_{rel} = 1$.

\section{Redirection in Online Therapy}
\label{sec:results}
We employ our formalism to study \redirection in online text-based therapies.
We first validate that our \redirection measure corresponds to human intuition in the therapy setting.
We then explore how \redirection relates to the development of the therapeutic relationship.
We move on to study the relation between \redirection and the patient's perception of the quality of the therapy. 
Finally, we apply the related measures discussed in Section~\ref{sec:method} to gain additional insight into which interactional dynamics might account for the observed trends.

\subsection{Validation studies}

\xhdr{Human validation} To check how aligned our measure is with our intuitive understanding of \redirection in this particular online therapy setting, we designed a redirection identification experiment. 
A participant is shown a pair of two short interactions, each consisting of 4 utterances (like the examples in \autoref{fig:intro_example}).  
Their task is to pick which one of them has a higher redirection effect in the second to last utterance (e.g., $T_k$ in \autoref{fig:intro_example}).

Each pair for this experiment is constructed by first picking a random therapy, selecting the utterance with the highest and lowest redirection effect according to our measure, and considering the interaction surrounding those utterances.
We select 10 such pairs using our redirection measure, and for comparison, we select 10 more pairs for each of the other measures discussed in Section \ref{sec:method}.
If a measure properly captures the redirection effect, we expect it to match the human rankings.

To respect the privacy of the data and adhere to IRB protocol, the task was performed by one of the authors who was authorized to read the therapy text. 
To avoid author bias, we administered the task in a blind fashion, with the participant not knowing which pairs were selected using which measure.

The participant ranking perfectly matched that of our redirection measure, compared to 8 out of 10 matches for the similarity difference, 5 out of 10 for orientation, and 4 out of 10 for uptake. 
While limited in scale due to the privacy restrictions on the data, these results suggest that our measure does indeed capture an intuitive notion of redirection which the other measures do not.
The experiment also inspired a qualitative analysis comparing examples in which the metrics disagree (Section \ref{sec:qualitative}).

\xhdr{Shuffle check} 
Given that \redirection is inherently an aspect of conversation flow, any proposed measure of it should be sensitive to utterance order.
To check for this intuitive property, we can {\em shuffle} a session's utterances:
we would expect no redirection to occur in this shuffled setting.

Indeed, while the average \redirection across all (non-shuffled) sessions is $6.40$, in the shuffled case it is near zero ($-0.025$; $p < 0.0001$ for the difference, Wilcoxon signed-rank test).

In our main analysis below, we  use the shuffle check to ensure that the trends we observe are actually related to conversational processes, as opposed to spurious non-conversational phenomena such as changes in session or utterance length.

\subsection{Evolution of the therapeutic relationship}
\begin{figure}[b]
    \centering
    \includegraphics[width=1\linewidth]{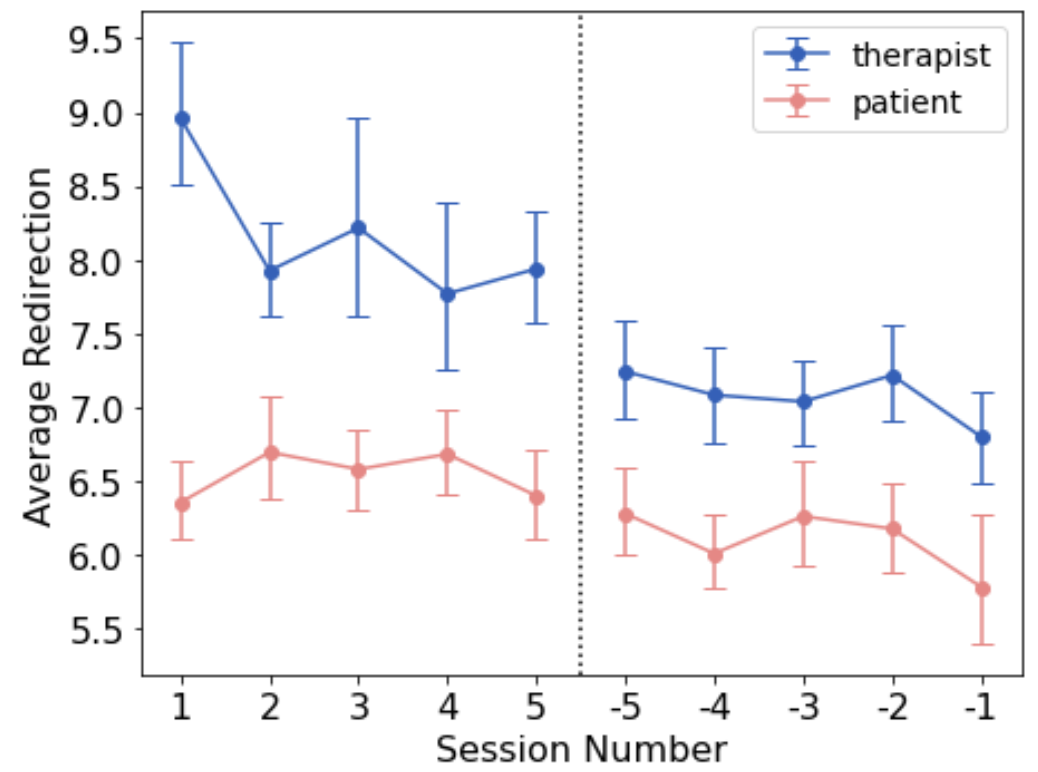}
    \caption{Average redirection across the first 5 and last 5 sessions. Throughout, error bars indicate 95\% confidence intervals estimated through bootstrap sampling.
    }
    \label{fig:avg_redirect}
\end{figure}

We now present our main analysis:  connecting \redirection to the development and quality of the therapeutic relationship.
We start with examining the change in \redirection as the therapist-patient relation develops, focusing on sustained therapies having at least $10$ sessions ($3,764$ in total).

\autoref{fig:avg_redirect} shows the average \redirection measure $T_{avg}$ and $C_{avg}$ in each session for the first 5 and last 5 sessions.
For both the patient and therapist, there is a downward trend in local \redirection as therapy progresses ($p < 0.0001$ according to a Wilcoxon signed-rank test comparing the first and last 5 sessions).
Thus, as the therapy progresses, redirection in conversation occurs less from both speakers, perhaps suggesting a smoother flow of the conversation with more stable conversational goals (see Section \ref{sec:qualitative} for a qualitative analysis).

\begin{figure}[t]
    \centering
    \includegraphics[width=1\linewidth]{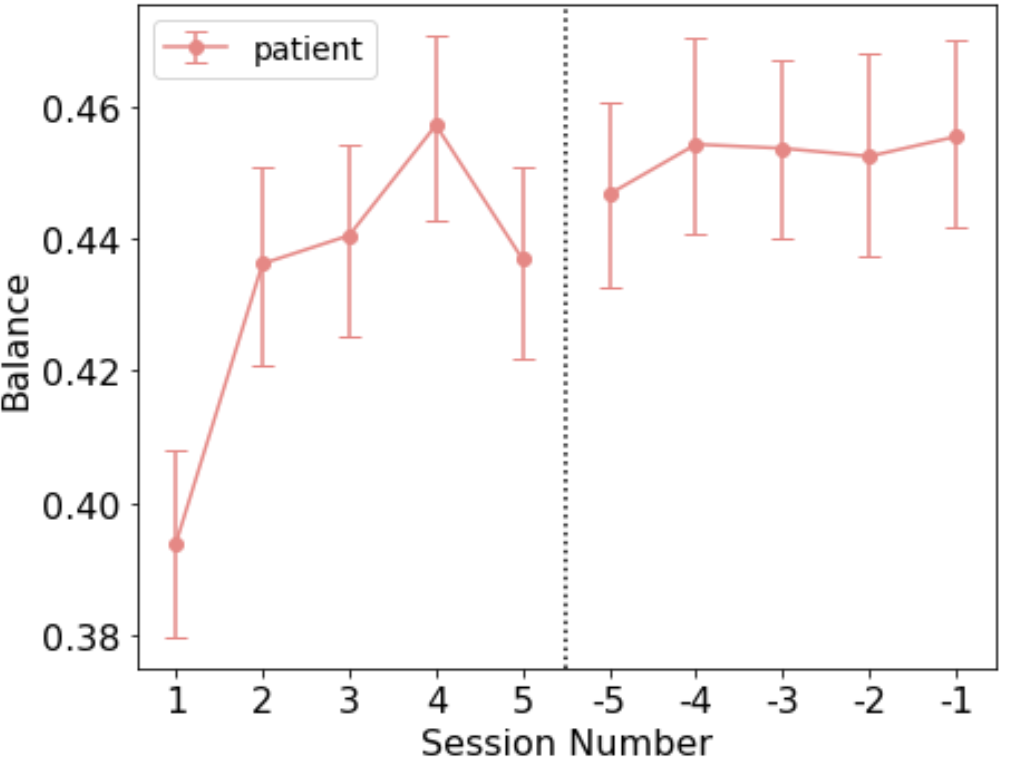}
    \caption{
    Patient \redirection relative to that of the therapist across the first 5 and last 5 sessions.
    }
    \label{fig:patient_rel_redirect}
\end{figure}

We now switch from considering each speaker's \redirection separately to examining how the \textit{balance} between the two evolves over time. 
\autoref{fig:patient_rel_redirect} shows the patient's relative \redirection $C_{rel}$ across the first 5 vs. last 5 sessions.
As the relationship progresses, the patient share of  \redirects increases (p < 0.0001), suggesting that the patients gain more relative control over the flow of the conversation.

We re-analyze the trends after shuffling the utterances within a session.
After the shuffle, all observed trends in the original data disappear (no statistical difference), suggesting that they are indeed tied to the conversation dynamics.

\subsection{Unsuccessful relationships}
We now investigate how \redirection is related to the quality of the therapeutic relationship.
For this purpose, we consider a therapeutic relation to be ``unsuccessful'' if the patient eventually abandons it via a request to switch therapists or to cancel the subscription \textit{and} provides a reason that explicitly cites dissatisfaction with the therapist or with the relation (see Appendix \ref{sec:appendixunsuccessfulrel} for a list of such reasons).
After filtering out therapies that had fewer than 3 sessions (to ensure that the patient actually ``gave it a try''), we are left with $817$ such unsuccessful relations.
For a meaningful comparison, we consider a control group with the same number of therapies where the patient did not request to cancel or switch the therapist, and that also has at least 3 sessions.
To discard effects due to the duration of the therapy and focus on signals that could be perceived early in the therapeutic process, we only consider the first 3 sessions in this analysis. 
\autoref{fig:good_bad_client} shows that in unsuccessful relationships, patients \redirect the conversations less than in the control group ($p = 0.018$; Mann-Whitney U test), while we do not find any difference for therapists ($p = 0.5$).
We note these differences disappear after shuffling the order of the utterances in the session ($p = 0.17$), thus they reflect differences tied to conversation dynamics.

\newcommand{\original}{actual\xspace}
\newcommand{\Original}{Actual\xspace}
\begin{table}
\centering
\begin{tabular}{lll|l}
\hline
\text{} & \text{\em Unsuccessful} & \text{\em Control} & \text{\bf  p-Value} \\
\hline
\text{\Original} & 6.06 & 6.91  & 0.018\textcolor{red}{*} \\
\text{Shuffled} & 0.14 & 0.043  & 0.17 \\
\hline
\end{tabular}
\caption{
\label{fig:good_bad_client}
Patient average \redirection is smaller at the start of (eventually) unsuccessful relationships (statistical significance indicated with \textcolor{red}{*}; p < 0.05). 
This effect disappears in the shuffled setting, as shown by the lack of statistical significance marker.
}
\end{table}

\subsection{Comparison to other measures}
\label{sec:appendixrelmeasuresresults}

\begin{table*}
\centering 
\begin{tabular}{p{6cm}p{2cm}p{2cm}p{2cm}p{2cm}}
\toprule
{\em Analysis} & {\em Orientation} & {\em Similarity Difference} & {\em Uptake}  & \textbf{{\em \Redirection}}\\ 
\midrule
Start/end difference in average value & - / $\downarrow$  & $\downarrow$ / $\downarrow$ & $\uparrow$ / $\uparrow$  & $\downarrow$ / $\downarrow$ \\
\midrule
Start/end difference in balance& $\uparrow$ / $\downarrow$ & - / - & - / -   & $\uparrow$ / $\downarrow$  \\
\midrule
Reflects temporal order (i.e., does shuffling remove the trend) & No & {\bf Yes} & No & {\bf Yes} \\
\midrule
Distinguishes unsuccessful relations & No & No & No & {\bf Yes}\\
\bottomrule 
\end{tabular} 
\caption{\label{citation-guide} 
Comparing \redirection with three related measures.
Slashes (``/'') separate  patient from  therapist effects. $\uparrow$~indicates a higher level in the end than the start; $\downarrow$ indicates vice-versa; ``-'' indicates no significance. 
}
\label{tab:measure_table}
\end{table*}

We also explore how our \redirection measure compares to the other related measures discussed in Section~\ref{sec:method}. 
The results are summarized in \autoref{tab:measure_table}. %
It is worth noting that none of the related measures show a significant difference between the ``unsuccessful'' therapies and the control group, and only the similarity difference measure exhibits temporal trends that pass the shuffling test.

We highlight here the results for orientation, since by capturing the redirection intent separately, they provide further context for interpreting the results discussed above.
Orientation shows a significant downward trend for therapists; thus the observed decrease in \redirection can be at least partially attributed to the decrease in their attempts to change the course of the conversation (rather than to the patient's unwillingness to realize those attempts).
Furthermore, there is no difference in patient orientation between the ``unsuccessful'' and control groups ($p=0.5$), suggesting that reduced patient redirection is not due to lack of patient \redirection bids, but rather because the therapist is not accepting those bids.

\section{Qualitative Analysis}
\label{sec:qualitative}

We conduct a qualitative analysis of high- and low-redirection examples to explore therapy strategies that are tied to this phenomenon and to further interpret the observed trends.
Additionally, we analyze examples where the redirection measure deviates from the related measures to check our intuition about differences in what they capture. 
The discussion follows examples from \autoref{tab:qualitative_ex}.

\newcommand{\exwidth}{11.5cm}
\newcommand{\colwidth}{.4cm}

\begin{table*}[!ht]
    \centering
    \small
    \begin{tabular}{p{0.2cm}Slp{\colwidth}p{\colwidth}p{\colwidth}p{\colwidth}}
        \toprule
        \textbf{\em \#} &\textbf{\em Example} & \textbf{\em R} & \textbf{\em O} & \textbf{\em SD} & \textbf{\em U} \\
        \midrule
        \textbf{1} &
        \parbox{\exwidth}
        {
        T: It seems like this anxiety doesn’t arise when you’re in a relationship with someone.\\
        P: Now that I remember no. I feel more secure.\\
        \textcolor{therapist}{\textbf{T: Where does the strong reaction to being alone originate from?}}\\
        P: Honestly, I think it has to do with my relationship with my family.
        } & High & Low & Low & Mid \\
        \hline 
        \textbf{2} &
        \parbox{\exwidth}{
        T: It seems like the issue is your lack of control over being alone.\\
        P: Yes, that’s interesting; I think that’s a part of it.\\
        \textcolor{therapist}{\textbf{T: Try sitting with your feelings and exploring them without blocking them, and see what insights come up for you.\\
        }}P: Okay, I’ll meditate on that. Thanks.
        } & High & Mid & Low & Mid \\
        \midrule 
        \textbf{3} &
        \parbox{\exwidth}{
        T: I believe using conflict resolution methods might help you better communicate with your husband about this issue, rather than withdrawing and shutting down. \\
        P: He keeps saying he doesn't want to get involved with her problems.\\
        \textcolor{therapist}{\textbf{T: Then, are you planning to talk to your daughter first, or are you waiting to speak with her counselor? \\
        }}P: I'll just ground her and take things away.
        } & Low & High & Low & Mid \\
        \midrule 
        \textbf{4} &
        \parbox{\exwidth}{
        T: When you are ready to process the trauma, we can go through it.\\
        P: Now, I’m feeling better mentally, but physically my heart is thumping, and I began having panic attacks at work.\\
        \textcolor{therapist}{\textbf{T: That sounds awful. Is it a thumping sensation in your chest or feeling in your ears?\\
        }}P: I wore a heart monitor in my chest. I feel a thump each time before my heart skips beats.}
        & Low & Low & High & High \\
        \midrule
        \textbf{5} & 
        \parbox{\exwidth}{
        P: I dislike how I get easily attached to a man I like. I had thoughts of being in a relationship just because we were spending time with each other.\\
        T: Some people are attachers in relationships, while some people it takes a while for the heart to thaw out.\\
        \textcolor{client}{\textbf{P: I feel like it may be because of not having my dad present.\\
        }}T: I think that is a good insight on your part. Your upbringing and personality formation definitely affects attachment styles.
        } & High & Low & Low & Low \\
        \midrule
        \textbf{6} & 
        \parbox{\exwidth}
        {P: How was your day?\\
        T: I’m also waiting for warm weather; I think today’s supposed to be pretty nice! How are things going at home for you?\\
        \textcolor{client}{\textbf{P: It’s going okay. I’m nervous about having to hit numbers at work every day though. That is going to be a challenge.\\
        }}T: I understand productivity requirements. It sounds like you have an amazing work ethic and your trainer is already impressed with you though.
        } & High & High & Mid & Low \\
        \midrule 
        \textbf{7} &
        \parbox{\exwidth}{
        P: I’m going to take a walk tonight if weather permits.\\
        T: The walk sounds like a great idea. It is good to let it all out, so don’t be afraid to vent as much as you need to.\\
        \textcolor{client}{\textbf{P: Thanks. Not used to having someone to vent to.\\
        }}T: Hope the rest of the day went better and the weather allows you to take a walk. Feel free to vent as things come up.
        } & Low & Low & Low & High \\
        \midrule
        \textbf{8} &
        \parbox{\exwidth}{
        P: When I have trouble concentrating, I tend to eat.\\
        T: Does that help you concentrate better?\\
        \textcolor{client}{\textbf{P: I don't know.\\
        }}T: So what’s behind eating when you are having trouble concentrating?
        } & Low & Low & Low & Mid \\
        \bottomrule
    \end{tabular}
    \caption{
    Qualitative examples comparing \RedirectionBoldLetter with related measures {\bf O}rientation, {\bf S}imilarity {\bf D}ifference, and {\bf U}ptake, all applied  to the colored utterance (\textcolor{therapist}{\underline{T}herapist} for the first four examples, \textcolor{client}{\underline{P}atient} for the remaining four).  Low/High/Mid: the bottom 25th percentile of the measure values for the respective speaker type, the top 75th percentile, and 25-75th percentile, respectively. Examples are paraphrased to preserve privacy.
    }
    \label{tab:qualitative_ex}
\end{table*}

\xhdr{\Redirection strategies}
In our setting, highly \redirecting therapist utterances typically involve exploration and surveying.
Therapists tend to use open-ended questions to assess their patients' given situations and viewpoints (Example 1: \textit{"Where does this [issue] originate from"}).
They may also suggest different perspectives or provide guidance on addressing these issues (Example 2: \textit{"Try sitting with your feelings ..."}). 
Conversely, less-\redirecting utterances tend to echo or validate the patient's statements or empathize with their struggles.
They may also reformulate what the patient has said and keep the focus on the immediate problem at hand (Example 4: \textit{"Sounds awful. Is it...")}.

For patients, highly \redirecting utterances tend to introduce personal experiences or feelings that shape the immediate course of the conversation.
They may initiate subjects they wish to explore or verbalize the specific obstacles they are facing (Example 6: Challenges at work).
Low-\redirecting patient utterances, on the other hand, extend the current subject of discussion.

These observations are consistent with the decreasing trend in redirection for both speakers.
Initially, patients and therapists both focus on introducing the context of the therapy and setting therapeutic goals. 
Therapists tend to actively discuss the objectives of the therapy and suggest behavioral strategies and cognitive techniques.
Patients, likewise, will share their background or expectations as they start out their therapeutic relationship.
Later sessions, conversely, usually consist of more concentrated discussions on the identified issues.

\xhdr{Comparison with other metrics}
We also examined cases where the \redirection measure is at odds with the related measures.
As expected, orientation can be high even when \redirection is resisted.
For instance, in Example 3, the therapist attempts to steer the conversation towards discussing who to talk to, but the patient disregards their suggestions and continues explaining their plans, thus causing \redirection to be low.

Importantly, unlike similarity difference, our metric captures utterances with \redirection effects even when the reply is not semantically similar.
In Example 2, the therapist \redirects the focus of the conversation by suggesting a potential solution to the patient's problem, which the patient acknowledges in their reply.
While similarity difference is low since acknowledging the suggestion is not semantically similar, the patient does reflect on the therapist's suggestion, which deviates from their prior discussion focused on the 
problem.

Our measure also differs from uptake in that it uses a reference point: the previous conversational context.
Example 7 illustrates a segment where the patient reflects on having someone to vent to. 
The therapist addresses the patient's utterance by continuing to encourage the patient to vent, exhibiting high uptake.
However, the focus of the conversation does not change, indicating low \redirection.

\section{Further Related Work}
\label{sec:related}
Our work relates to prior research on conversational dynamics, analyzing the development and quality of therapeutic relationships, and exploring language on \mh platforms.

\xhdr{Development and quality of therapeutic relationships}
The development of the patient-therapist relationship and its impact on therapy outcomes have been extensively studied \cite{gelso_relationship_1985, norcross_therapeutic_2010}.
Previous work characterized therapeutic alliance \cite{goldberg_machine_2020} and ruptures \cite{tsakalidis_automatic_2021} through emotional engagement \cite{christian_assessing_2021}, sentiment \cite{syzdek_client_2020}, linguistic coordination \cite{nasir_modeling_2019}, and synchrony \cite{dore_linguistic_2018}.
Our work characterizes an additional dimension of the patient-therapist relationship based on the joint act of redirection.

\xhdr{Conversational dynamics}
Prior computational work examined different aspects of conversation flow \cite{zhang_conversational_2016}, including work on topic segmentation \cite{eisenstein_bayesian_2008,nguyen_modeling_2014,purver_topic_2011,glavas_two-level_2020,jiang_superdialseg_2023} and topic shift \cite{xie_tiage_2021}.
While these conversation-level concepts are related, our probabilistic measure seeks to quantify the immediate effect of a single utterance, an effect different from semantic similarity and topic coherence.
Furthermore, we develop a framework for a \textit{longitudinal} analysis of conversational dynamics in the distinctive domain of online \mht.

\xhdr{Online \mh platforms}
Prior literature extensively explores linguistic behaviors and conversational choices users make in online \mh-related platforms, whether in crisis counseling platforms \cite{althoff_large-scale_2016, zhang_balancing_2020}, peer-support platforms \cite{yang_self-disclosure_2017,pruksachatkun_moments_2019,yang_modeling_2024}, or therapy platforms \cite{malgaroli_association_2023}.
Accompanying growing attention to online \mh resources and use of technology in psychotherapy \cite{anthony_use_2003, barak_comprehensive_2008}, many studies highlighted the benefits of online social support \cite{de_choudhury_language_2017, newman_its_2011}.
We specifically focus on \textit{long-term} sustained relationships in text-based therapy platforms,
offering a novel perspective on development of the patient-therapy relation through longitudinal analyses.

\section{Discussion and Conclusion}
\label{sec:discussion}
\Redirection is a joint act often performed by therapists and patients.
Its realization requires both speakers to understand the initial direction of the conversation, one's attempt to \redirect it, and the other's compliance with this attempt. 
As such, its study has the potential to characterize the patient-therapist dyad, in particular with respect to their ability to negotiate the focus of the conversation.

In this work, we introduce a computational method for quantifying the \redirection effect of an utterance and apply it to a \mh-therapy domain.
We find that both the patient and the therapist \redirect less as therapy progresses.
Our qualitative analysis suggests that after initial exploration and contextualization prompted by both speakers, the relationship matures to a more stable stage with less \redirection.
Moreover, we reveal that
the less patients \redirect early on, the more likely they are to 
eventually express dissatisfaction with the therapist and abandon the relationship.

\xhdr{Connection with psychotherapy literature} Several decades of research in psychotherapy suggest that the ``treatment model'' approach in which an expert clinician provides curative treatment that the patient passively receives is not well supported by outcomes or engagement data \cite{bohart_client_2000,duncan_clients_2000}. 
Instead, shifting focus from which aspects of treatment are delivered to how dyads collaboratively negotiate therapeutic conversations has provided evidence of improved access, efficiency, and effectiveness of \mh services for ``patient-led'' approaches \cite{carey_will_2010,carey_effective_2013,huber_agency_2021}. 
Our results
offer a computational perspective into one dimension of patient agency in 
therapeutic relationships.  

\xhdr{Future work} A computational approach to \redirection enables us to observe the evolution of therapeutic relationships and could assist therapists in fostering them. 
A complementary line of work can include a more in-depth exploration of how to foster healthy levels of redirection in therapy and the causal relationship between \redirection and the quality of therapeutic relationships.
For instance, labeling messages with therapeutic strategies and techniques can provide insight into the effectiveness of each strategy and determine which ones are more applicable for specific contexts (e.g., when patients resist therapists' \redirection attempts).

In their current observational form, these results suggest that early conversational patterns can signal the eventual dissatisfaction of the patient.
Future work could examine the predictive power of these ties and test the extent to which they might be explained by other (unalterable) factors, such as the characteristics of the patient or of their condition.

Our methodology can extend beyond the \mh domain and be applied to other conversation-rich domains, such as education, online discussions, or interviews, where discussions are carried out in relatively unstructured ways with only a few general agendas set.
Exploration of \redirection in these contexts may present a unique perspective into how speakers are able to \redirect and control the flow of the discussions.

Finally, understanding how humans redirect the flow of conversations is important for supporting more naturalistic human-AI conversations, where the AI could pick up on humans' redirection attempts and initiate their own.

\section{Limitations}
\label{sec:limitations}

In the \mh context, how to define successful therapeutic relations is always an open question.
Our work uses patient-provided reasons for canceling the therapy or switching therapists that explicitly mention dissatisfaction with the therapist as an imperfect indicator for failure of the therapeutic relationship. 
A reliable positive signal, on the other hand, is difficult to define, especially in a therapy domain where patients can have hidden agendas, present misleading information, or choose to remain in treatment for a number of reasons \cite{newman_when_2003}.  
We thus do not consider the control group to necessarily contain ``successful'' relationships, and thus the interpretation of the results of the comparison should be interpreted accordingly. 

Furthermore, these results are purely observational, and future work would be needed to establish if intervening to change the amount of realized redirection (e.g., through therapist training) would have an effect on the quality of the therapeutic relationship.

The notion of session we employed captures the structure of therapy that arises as patients and therapists maintain a long term relationship spanning multiple conversations, rather than just a single encounter.
However, sessions' characteristics change as time progresses.
We observed that the session length and the token count of utterances in sessions both decrease. 
While our shuffling experiments suggest that our observations are not merely a result of this variability and are tied to the actual dynamics of the conversation, further work is needed to fully account for this variability.

Our work focuses on text-based conversations.
While these are a substantial component of online therapy, they do not capture the occasional video conversations that patients and therapists might have.  
Future work could explore video conversations and the role they have in the development of relationships (although additional de-identification challenges arise when working with video data).

As our \redirection measures reflect change in the subject of the conversation, they require the reply of the utterance to be able to be calculated. 
This constraint can be restrictive in practice, as an automated system would be unable to analyze an ongoing conversation without a reply it hasn't received yet. 
Exploring how we can predict whether the upcoming reply will cooperate with one's attempt to redirect remains an interesting direction for future work.

\xhdr{Ethical Concerns} 
As our work involves highly sensitive data in the form of patient chat-therapy history and surveys, all personally identifiable information was removed. 
Access to the data was strictly limited to the authors of the paper, and data was processed on restricted servers and will be removed upon publication.
All large language models trained on the data were strictly internal and discarded after the analysis.

This work is done in close collaboration with the therapy platform Talkspace. 
All participants consent to the use of their data in aggregate and de-identified format for research purposes as part of the Terms of Service they review during onboarding (in the case of the patients)  or as part of information reviews done during the hiring process (in the case of the therapists). 
All individuals may opt out of the use of their data for research purposes at any time without penalty. 

Given the sensitivity of the domain and the limitations discussed above, extensive work is still required before insights from our results could be used in patient-facing systems.

\section*{Acknowledgements}
We thank Liye Fu, Emily Tseng, Khonzoda Umarova, and Tony Wang for initial pre-processing of the data. 
We are grateful for insightful discussions with Team Zissou---including Jonathan P. Chang, Yash Chatha, Nicholas Chernogor, Tushaar Gangavarapu, Kassandra Jordan, Seoyeon Julie Jeong, Ethan Xia, and Sean Zhang---and for the feedback we received from the anonymous reviewers.
Finally, we would like to express our gratitude to the patients and therapists on Talkspace who generously shared their experiences and therapy data for research purposes.
This material is based upon work supported in part by the U.S. National Science Foundation under Grant No. IIS-1750615 (CAREER).
Any opinions, findings, and conclusions or recommendations expressed in this material are those of the author(s) and do not necessarily reflect the views of
Cornell University, the National Science Foundation, or Talkspace.

\bibliography{refs}

\appendix
\label{appendix:appendixsection}
\section{Additional Data Details}
\subsection{Session Split}\label{sec:appendixsessionsplit}
Adapting the methodology in \citet{kushner_bursts_2020}, we chose a value of $N$ for the session split based on the number of sessions it produces for $10,000$ random conversations. 
The median, mean, and standard deviation all plateau starting from $N = 50$ and level off around $N = 100$. Setting $N$ at $100$ will provide the reliability of the measure in relation to the selection of its value.
\begin{figure}[H]
    \centering
    \includegraphics[width=0.8\linewidth]{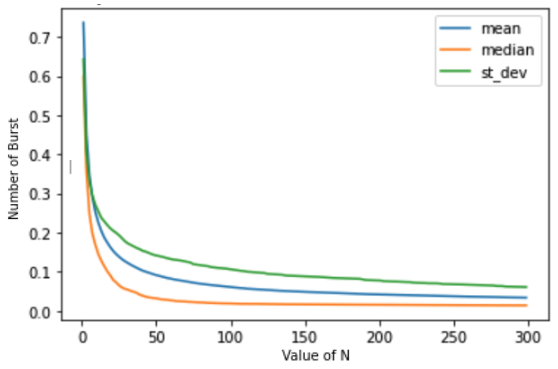}
    \caption{Number of sessions per $N$ from 10,000 random conversations. 
    }\label{fig:sessions}
    \end{figure}

\subsection{Switch / Cancel Surveys}
During therapy, patients may choose to switch therapists or cancel their plan by filling out surveys.
In the survey, they are asked to provide a reason for the switch/cancel, which they either select from a fixed list, or enter their own in free text.
The provided switch reasons include:
\begin{enumerate}
    \item[s1] I could not find a time to meet with my provider.
    \item[s2] I couldn't form a strong connection with my provider.
    \item[s3] I don't feel like my provider was responsive enough.
    \item[s4] I just want to try someone new.
    \item[s5] I want to select a provider with a different gender.
    \item[s6] I was unsatisfied with the quality of care.
    \item[s7] Disabled.
    \item[s8] Dissatisfied with app.
    \item[s9] Dissatisfied with provider.
    \item[s10] Expensive. 
\end{enumerate}

The provided cancel reasons include:
\begin{enumerate}
   \item[c1] I met my goal / I feel better.
   \item[c2] The cost doesn't fit my budget.
   \item[c3] The treatment provided by my therapist was not helpful.
   \item[c4] I had technical issues.
   \item[c5] My therapist was not responsive to my messages.
\end{enumerate}

\subsection{Defining Unsuccessful Relations}\label{sec:appendixunsuccessfulrel}
The dataset of ``unsuccessful'' therapies with the patient abandoning the relationship includes therapies where the patient requested either a switch with reasons s2, s3, s4, s6 or a cancellation with reason c3.

\section{Operationalization}

\subsection{Implementation Details}\label{sec:appendiximplementation}
For our \redirection model, we fine-tuned Gemma-2B 
\cite{mesnard_gemma_2024} with 4-bit QLoRA \cite{dettmers_qlora_2023} using the huggingface library.
We use a 90/10 split for training and validation, and trained for $2$ epochs, with LoRA rank = $16$ and dropout = $0.05$, context length $4096$, batch size $2$, learning rate 2e-4, AdamW optimizer \cite{loshchilov_decoupled_2017}, achieving a validation perplexity of $10.97$. 
The total train time is approximately $43$ hours on 2 NVIDIA RTX A6000 GPUs.
For the model experiments, we conducted a hyperparameter search over learning rates [2e-5, 2e-4] and LoRA rank = [8, 16, 64], and used fixed values for the rest of the parameters.

We used the ConvoKit Python package \cite{chang_convokit_2020} to calculate orientation to employ the same methodology outlined in \citet{zhang_balancing_2020}.
We trained two separate models for therapist and patient orientation, using dependency-parse arcs representations for both speakers and 12 SVD dimensions.

To calculate uptake, we fine-tuned a pre-trained BERT-base model for next utterance classification. 
Two separate models were trained: one for predicting patient's utterance after a therapist and the other for vice-versa.
Our training setup follows the original paper \cite{demszky_measuring_2021}.
We fine-tune our model for 1 epoch with a batch size of 16, max length of 512 tokens for patient's and therapist's utterance each.
The learning rate is set at 6.24e-5, with linear decay and AdamW optimizer \cite{loshchilov_decoupled_2017}
The total train time is approximately 12 hours for each model on 2 NVIDIA RTX A6000 GPUs. 
We used the two models with the original source code from \cite{demszky_measuring_2021} to calculate uptake.

For similarity difference, we used a pre-trained sentence BERT model 'multi-qa-MiniLM-L6-cos-v1' to map utterances into a 384 dimensional dense vector space. We calculated the similarity between two utterances using cosine similarity of their embeddings.
\subsection{Used Artifacts}\label{sec:appendixartifacts}
We list the following artifacts and their licenses that are used in the work. 
\begin{itemize}[leftmargin=*]
    \item ConvoKit 2.5.3:
    \\ \url{https://convokit.cornell.edu/}, MIT License
    \item PyTorch 2.2.1:
    \\ \url{https://pytorch.org}, BSD-3 License 
    \item Sentence Transformers 3.0.0:
    \\\url{https://github.com/UKPLab/sentence-transformers}, Apache License 2.0
    \item Transformers 4.38.2:
    \\ \url{https://github.com/huggingface/transformers}, Apache License 2.0
    \item Conversational Uptake Source Code:
    \\ \url{https://github.com/ddemszky/conversational-uptake}, MIT License
\end{itemize}

\section{Additional Application to US Supreme Court Oral Arguments}
\label{sec:supreme}
We also examine how our \redirection framework can be applied in other domains in addition to mental health.
In particular, we apply our method to a publicly available dataset of U.S. Supreme Court oral arguments \cite{danescu-niculescu-mizil_echoes_2012, chang_convokit_2020}.
Although court proceedings differ from therapy in terms of topics, goals, and interaction styles, their relatively unstructured and dynamic nature enables an initial exploration of how such discussions are redirected.

In this setting, we focus on the interactions between justices and lawyers.
The power dynamics between these distinct roles reflect the asymmetric relationship between therapists and patients in \mh domains, where one party generally holds more influence over the conversation.

As expected, our analysis reveals that justices redirect the conversation significantly more than lawyers ($p < 0.001$, according to a Wilcoxon signed-rank test).
Our findings suggest that justices exercise more control over the flow of the discussion, steering it towards issues they consider critical to the case.

To further examine how \redirection unfolds in these exchanges, we use a Bayesian distinguishing word analysis, ``Fightin' Words'' \cite{monroe_fightin_2017} to compare high and low \redirection phrases from both speakers.
For justices, highly \redirecting utterances frequently involve assertive questioning ("may ask", "ask you", "if he", "is this") or references that draw the court's attention to specific matters ("the court", "court of"). 
In contrast, low \redirecting utterances from justices tend to include responses ("all right", "that right", "no no") or clarifications ("you mean", "mean that") of subjects raised by other parties.

Conversely, highly \redirecting utterances by lawyers often direct the court's ("the federal", "this court", "this state") focus to new arguments or highlight their perspectives on the case ("it seems", "to me", "that the"). 
Low lawyer \redirecting utterances, however, tend to affirm and acknowledge the justice's statements ("yes sir", "your honor", "oh yes") or continue discussing the current issue at hand ("it was", "it for").

For reference, we provide additional examples of high and low \redirection from both speakers in \autoref{tab:supreme_court_table}.

\newcommand{\suexamplewidth}{13cm}
\begin{table*}[htbp]
    \centering
    \small
    \begin{tabular}{p{0.5cm}Slp{1.5cm}}
        \toprule
        \textbf{\em \#}& \textbf{\em Example} &\textbf{\em Redirection} \\
        \hline
        \textbf{1} &
        \parbox{\suexamplewidth}{
        J: That's on the outer, outer belt. \\
        L: On the outer, outer belt. Now there's no dispute about it. It was admitted by the President of Santa Fe so that the evidence is here but the Commission simply ignored it. \\
        \textcolor{therapist}{\textbf{J: Let me ask you [...] Are the gateways of Danville, Decatur Springfield along the Wabash which is a wholly owned subsidiary of the Pennsylvania, are they of any consequence? \\
        }}L: The Decatur gateway is because as Your Honor would see that leads through the Wabash to the Hannibal bridge crossing the Mississippi and then on to Kansas City. [...] The Springfield gateway is an important gateway but of lesser importance [...].
        } & High \\
        \hline 

        \textbf{2} &
        \parbox{\suexamplewidth}{
        J: You happen to know what was your practice, if you had a practice of having you're statements to the grand jury in summary or explanation of what the evidence disclosed or manifest, was that taken down by a stenographer? \\
        L: No, I don't recall it ever being done [...]. \\
        \textcolor{therapist}{\textbf{J: Did Heras adjure the conspiracy or did you just not have enough -- have recent acts on his part? \\
        }}L: Well, in all frankness, I don't think that Heras have adjured the conspiracy [...].
        } & High \\
        \hline 
        
        \textbf{3} &
        \parbox{\suexamplewidth}{
        J: [...] This protection by the McCarran Act offer the individual State, protection to the State from the paramount federal power is difficult to reconcile with the theory after making one State subject to the laws of another State, in which laws they have no part in making. Do you subscribe to that language? \\
        L: Well, the -- latter part would raise a question of possible [...]. \\
        \textcolor{therapist}{\textbf{J: Well it didn't make any sense to me at all.\\
        }}L: Well -- I said the -- the latter part may raise the question as to, the possible conflict were Nebraska interprets deceptive practice in one way -- Nebraska corporation [...].
        } & Low \\
        \hline 
        
        \textbf{4} &
        \parbox{\suexamplewidth}{
        J: [...] Do you think it would be a permissible reading of the Act to say that as far as conspiracy versus substantive counts, yes, Congress must be taken to have intended cumulative punishments, authorized cumulative punishments, but with respect to the two substantive accounts, it cannot be so taken to have intended? \\
        L: Well, I can't say that you can't read it that way [...] \\
        \textcolor{therapist}{\textbf{J: I think your view is to whether there's a difference between the approach that the Court should take in a case of this kind, if there is any difference.\\
        }}L: Well, I think the thing that bothers me is that in the context in which these cases arise, the question of power goes to the offense [...].
        } & Low \\
        \hline 

        \textbf{5} &
        \parbox{\suexamplewidth}{
        L: [...] I just wanted to say I thought that distinguished this case from many others which might be put. \\
        J: The question I wanted to ask you was, is the restaurant itself, either on its menus or its advertising literature, carry any notation that it's identified in any way with the Delaware [...] \\
        \textcolor{client}{\textbf{L: [...] The third point to emphasize is that, [...] the entire enterprise [is] bound up financially into a single project. The Supreme Court of Delaware, [...] ruled that the leases could be permitted only to the extent that such leasing is necessary and feasible to enable the Authority to finance the project [...]. \\
        }}J: [...] That third fact would apply to every leasing because whether the State leases in order to derive money from the leasing or part of the money for maintaining the state enterprise [...] it seems to be immaterial [...].
        } & High \\
        \hline

        \textbf{6} &
        \parbox{\suexamplewidth}{
        L: [...] that I would doubt whether it was a deep-seated. \\
        J: Look at the alibi of Mapp for a good illustration. \\
        \textcolor{client}{\textbf{L: [...] one of the parts that bothers me here, Mr. Justice, it's the converse in a sense of what I said earlier. I spoke of the affirmative effect of a decree pointing out the constitutional duty of the legislature. [...] you said since the Supreme Court of Tennessee refused to act, that established that there was no violation of the Tennessee Constitution. \\
        }}J: That isn't what I said. I said that that decision isn't done nothing [...].
        } & High \\
        \hline 

        \textbf{7} &
        \parbox{\suexamplewidth}{
        L: [...] There would be no reason for him to have been walking out along that ledge of that barge [...]. \\
        J: If he had been ordered to go to do the work? \\
        \textcolor{client}{\textbf{L: And he had been ordered to go do that work on the raft, that's correct sir. \\
        }}J: And what you're saying is as a matter of law, it has to defend as a matter -- you have to defend it, right, it has been found as a matter of law. And when he walked from there -- from the Pfeifer or to the Winisook catwalk, and before he fell on it, he was at that time as a matter of law, no longer engaged in any duty as mate.
        } & Low \\
        \hline 

        \textbf{8} &
        \parbox{\suexamplewidth}{
        L: They are outlined in our briefs and needless to say, I do not have the time and shouldn't take the time to outline them all [...] let's make a further argument or conclusion or state my own opinion and then I shall develop the facts [...]. \\
        J: If you don't mind I suggest to you, I think you help us a lot more if you could guide us through what you consider to be the controlling thing instead of characterizing them all. \\
        \textcolor{client}{\textbf{L: Your Honor it's quite, I'm sure Your Honor is quite right. \\
        }}J: [...] Now I hope you are going to discuss them separately [...].
        } & Low \\
        \bottomrule

    \end{tabular}
    \caption{Examples comparing high and low \redirection from justices and lawyers in Supreme Court oral arguments. The measure is applied to the colored utterance (\textcolor{therapist}{\textbf{J}ustice} for the first 4 examples and \textcolor{client}{\textbf{L}awyer} for the remaining 4). Low indicates the bottom 25th percentile of the measure values for the respective speaker type; High indicates the top 75th percentile.}
    \label{tab:supreme_court_table}
\end{table*}

\end{document}